\documentclass[runningheads]{llncs}
\usepackage{graphicx}
\usepackage{amssymb}
\usepackage{amsmath}
\usepackage{multirow}
\usepackage{booktabs}
\usepackage{bbding}
\usepackage{subfigure}
\usepackage[colorlinks, citecolor=blue]{hyperref}
\usepackage{cleveref}
\usepackage{xcolor}
\Crefname{section}{Sec.}{Secs.}
\Crefname{table}{Tab.}{Tabs.}
\Crefname{figure}{Fig.}{Figs.}
\Crefname{equation}{Eq.}{Eqs.}

\begin{document}

\title{FITA: Fine-grained Image-Text Aligner for Radiology Report Generation}
%
%

\author{Honglong Yang \and HuiTang \and Xiaomeng Li\thanks{*Corresponding author}}

%
\institute{Department of Electronic and Computer Engineering, The Hong Kong University of Science and Technology \\ \email{honglong0216@gmail.com, eehtang@ust.hk, eexmli@ust.hk}}
\maketitle              
\begin{abstract} 

Radiology report generation aims to automatically generate detailed and coherent descriptive reports alongside radiology images. Previous work mainly focused on refining fine-grained image features or leveraging external knowledge. However, the precise alignment of fine-grained image features with corresponding text descriptions has not been considered. 
This paper presents a novel method called \textbf{Fine-grained Image-Text Aligner (FITA)} to construct fine-grained alignment for image and text features. It has three novel designs: \textbf{Image Feature Refiner (IFR)}, \textbf{Text Feature Refiner (TFR)} and \textbf{Contrastive Aligner (CA)}. 
IFR and TFR aim to learn fine-grained image and text features, respectively. We achieve this by leveraging saliency maps to effectively fuse symptoms with corresponding abnormal visual regions, and by utilizing a meticulously constructed triplet set for training.
Finally, CA module aligns fine-grained image and text features using contrastive loss for precise alignment.
Results show that our method surpasses existing methods on the widely used benchmark.




\keywords{Radiology Report Generation $\cdot$ Multimodal Alignment $\cdot$ Fine-Grained Representation}
\end{abstract}

\section{Introduction}
Radiology report generation aims to automatically generate free-text descriptions and diagnoses for clinical radiographs, thereby alleviating the burden on radiologists and providing guidance to inexperienced radiologists regarding potential abnormalities. In our study, a radiology report typically comprises findings that describe medical observations and impressions that summarize significant observations. This paper mainly focuses on developing an automated method for generating free-text reports from chest X-rays. 
\begin{figure}
    \centering
    \includegraphics[width=0.9\linewidth]{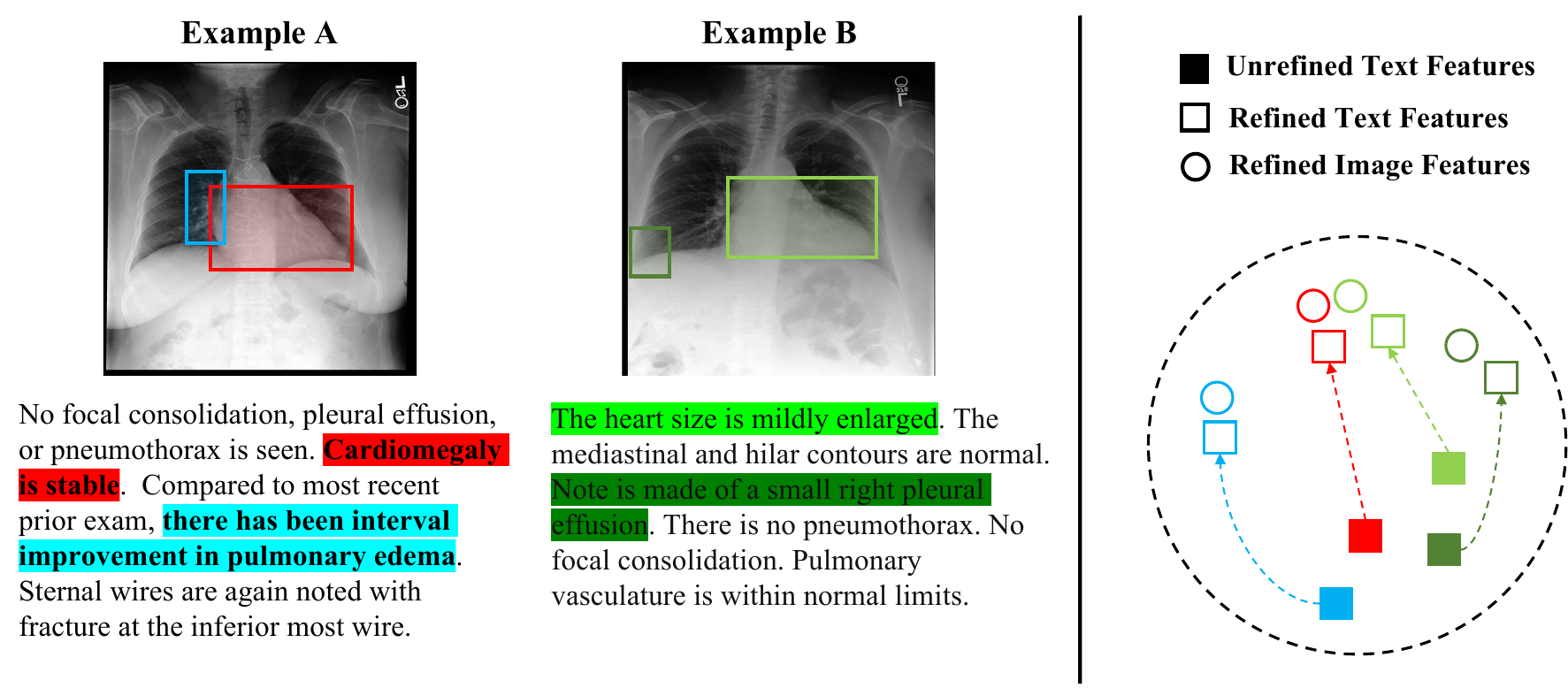}
    \caption{Left are two paired X-ray images and reports from MIMIC-CXR dataset. The right is the feature distribution for the pinpoint image patches and colored sentences. The circle denotes visual features and the rectangle denotes text features, with aligned visual and textual parts marking in the same color.}
    \label{fig:fine_grained_example}
    \vspace{-6mm}
\end{figure}

Early works~\cite{R2Gen,jing2017automatic,li2018hybrid,R2GenCMN,h_rnn} in radiology report generation followed the image captioning paradigm that adopted the encoder-decoder architecture. They primarily focused on prompting the network architectures to capture fine-grained image features and fuse image features and report features~\cite{R2Gen,jing2017automatic,R2GenCMN,you2021aligntransformer,cao2023mmtn,liu2021contrastive,wang2023metransformer}. Despite their ability to discern detailed image features, these methods struggled to produce high-quality radiology reports. Recent studies~\cite{zhang2020radiology,li2019knowledge,ppked,rgrg,li2023dynamic,jin2023promptmrg,huang2023kiut} incorporated external knowledge, such as pre-trained classification models~\cite{irvin2019chexpert,smit2020CheXBert,cohen2022torchxrayvision} and knowledge graphs~\cite{jain2021radgraph,zhang2020radiology}, to improve report generation. For example, KiUT~\cite{huang2023kiut} achieved expressive results by injecting visual and contextual information with external clinical knowledge that fuses the pre-defined clinical symptom graph, symptom probabilities from pre-trained classification model, and symptom embeddings. PromptMRG~\cite{jin2023promptmrg} generates high-quality reports by converting the diagnostic results from a disease classification branch to guide report generation. While these approaches have shown impressive performance, they primarily focus on exploring the fine-grained details of radiology images or incorporating various external knowledge, without considering the precise alignment between image patches and corresponding texts. For example, as shown in \Cref{fig:fine_grained_example}, a report may discuss multiple diseases with corresponding image patches marked in different colors, but previous methods have overlooked the distinctiveness between these descriptions, failing to accurately match them with the image patches. This challenge highlights the need for fine-grained alignment between refined image and text features to establish a direct correspondence between image patches and their detailed descriptions.
\par
To this end, we propose a novel framework called Fine-grained Image-Text Aligner (FITA), which consists of three modules: Image Feature Refiner (IFR), Text Feature Refiner (TFR) and Contrastive Aligner (CA). 
IFR focuses on extracting semantic features and identifying abnormal visual regions within radiological images. This is achieved by combining a classification loss and leveraging Grad-CAM~\cite{selvaraju2017grad}, derived from a pre-trained medical classification model~\cite{cohen2022torchxrayvision}.
TFR, on the other hand, aims to extract semantic features from textual reports while discerning subtle differences between abnormal and normal sentences. 
Finally, the CA module aligns the refined image and text features, ensuring consistency between the multi-modal representations.


In summary, our primary contributions can be outlined as follows: (1) We have innovatively proposed the FITA model with three modules: Image Feature Refiner (IFR), Text Feature Refiner (TFR) and Contrastive Aligner (CA) to capture both the fine-grained details within radiological data and alignment between refined images and textual descriptions. (2) We have introduced a novel approach that utilizes saliency maps to explicitly extract the abnormal visual regions within radiological images and the related symptoms. (3) Results on the widely used benchmark show that our method surpasses the performance of previous state-of-the-art methods.

\vspace{-2mm}
\section{Related Work}
Most works leveraged the encoder-decoder architecture as the image captioning~\cite{lu2017knowing,xu2015show} due to their similarity. However, radiology report generation poses greater challenges due to the longer report length and the fine-grained nature of radiology images and reports. Previous studies have employed various strategies to solve the challenges, broadly falling into two main categories. \\
\noindent \textbf{Prompting Model Structure.} Various methods tried to improve the model structure to extract better representations. AlignTransformer~\cite{you2021aligntransformer} proposed Align Hierarchical Attention (AHA) and Multi-Grained Transformer (MGT) to hierarchically align the visual regions and disease tags. Contrastive Attention~\cite{liu2021contrastive} innovated by introducing aggregate and differentiate attention to capture contrastive information between abnormal and normal images. METransformer~\cite{wang2023metransformer} introduced bilinear pooling module~\cite{lin2015bilinear} to capture high-order interactions between the input fine-grained feature. Besides, several works introduced the memory mechanism to remember the report's pattern. R2Gen~\cite{R2Gen} proposed relational memory module to record important information during the decoding process. MMTN~\cite{cao2023mmtn} proposed memory augment module in which key and value memory matrix are used to learn the relationship between two features. \\
\noindent \textbf{Introducing External Knowledge.} Other types of work explored injecting external knowledge such as disease tags~\cite{irvin2019chexpert,smit2020CheXBert}, retrieved reports and knowledge graph~\cite{jain2021radgraph} to assist report generation. PPKED~\cite{ppked} combined the abnormal findings, knowledge graph and retrieved reports to imitate the working patterns of radiologists. RGRG~\cite{rgrg} leveraged object detector as a region guidance for report generation, which enabled the decoder to explicitly leverage the disease information for generation. KiUT~\cite{huang2023kiut} achieved expressive results by integrating visual and contextual knowledge with external clinical insights, introducing an Injected Knowledge Distiller for knowledge injection in the final decoding phase. 

\section{Methodology}
\subsection{Problem Formulation}
\begin{figure}
    \centering
    \includegraphics[width=0.9\linewidth]{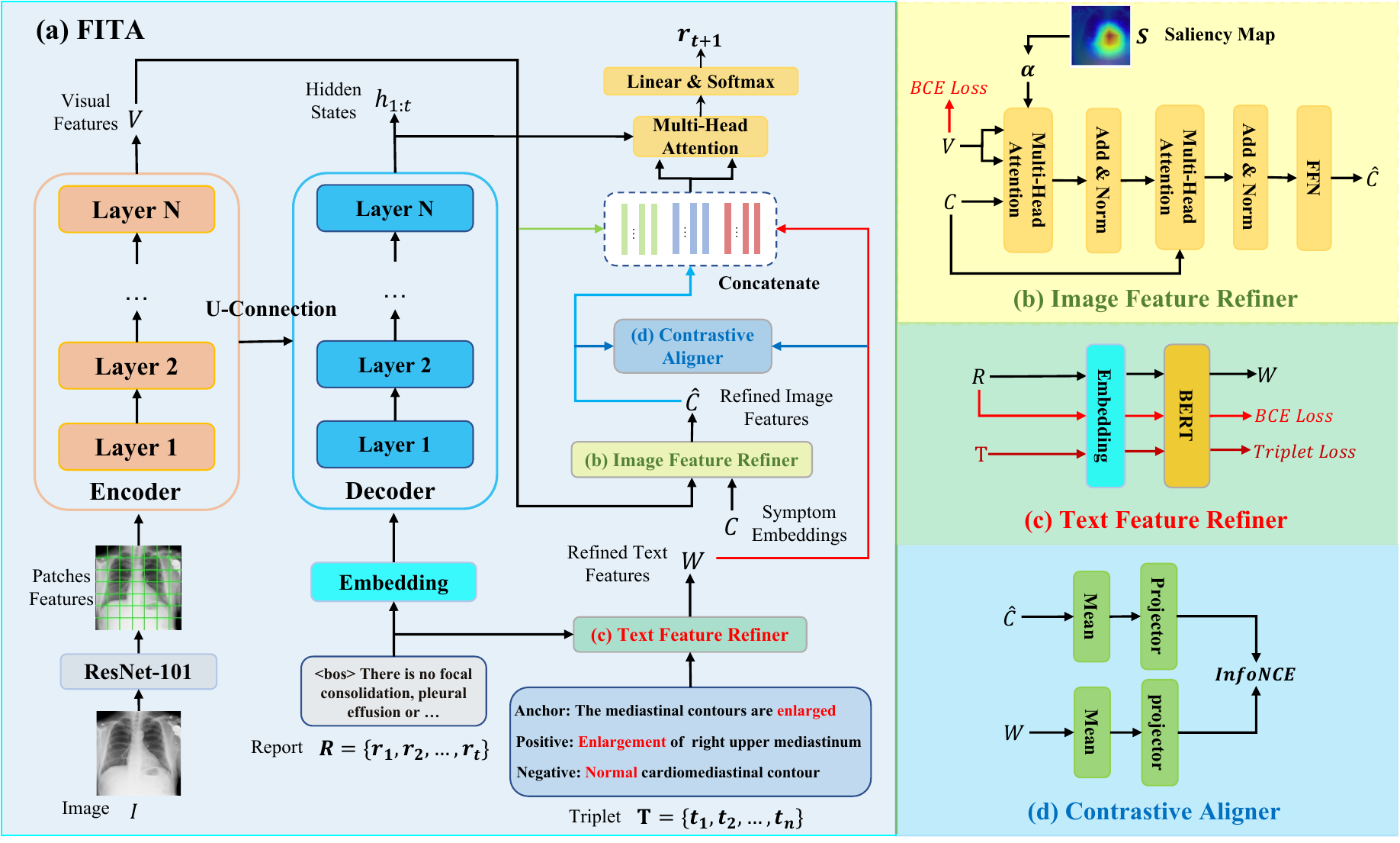}
    \caption{Illustration of our proposed Fine-grained Image-Text Aligner (FITA) approach, which adopts the U-Transformer structure and introduces three modules including Image Feature Refiner (IFR), Text Feature Refiner (TFR), Contrastive Aligner (CA).}
    \label{fig:overview}
    \vspace{-6mm}
\end{figure}

Given a radiology image $I$, the model is required to generate a descriptive radiology report $R=\{r_1, r_2,..., r_{N_R}\}$ where $r_i$ is the token for report and $N_R$ is report length. The recursive generation process can be formulated as $P(R|I)=\prod_{t=1} p(r_{t+1}|r_1,r_2,...r_t, I)$ and the model is trained to minimize
\begin{equation}
    \mathcal{L}_{CE}(\theta) = -\sum_{t=1}^{N_R}\log (p_{\theta}(r_{t+1} | r_{1:t})).
\end{equation}
Based on the above formulation, we propose a novel generation framework named Fine-grained Image-Text Aligner (FITA) (as the overview shown in \Cref{fig:overview}), to capture both the fine-grained details within radiological data and the crucial alignment between refined images and textual descriptions.
\subsection{FITA: Fine-Grained Image-Text Aligner}
As shown in \Cref{fig:overview} (a), Our FITA adopts the U-Transformer structure as KiUT~\cite{huang2023kiut} and proposes Image Feature Refiner(IFR), Text Feature Refiner(TFR) and Contrastive Aligner (CA) for fine-grained image-text alignment. Specifically, we use U-Transformer to extract visual features $V\in \mathbb{R}^{N_I\times d}$ and hidden state $h_{1:t}\in\mathbb{R}^{t\times d}$ where $N_I$ is the patches number and $t$ is the generation step. The IFR and TFR are utilized to refine image and text representations while CA is used to align the refined features. The whole process can be formulated as:
\begin{equation}
\begin{aligned}
    &\hat{C} = IFR(C, V, S), \ W_{1:t} = TFR(r_1, r_2,...,r_t), \\
    &r_{t+1} = Softmax(MHA(h_{1:t}, [V; \hat{C}; W_{1:t}])),
\end{aligned}
\end{equation}
where $S\in \mathbb{R}^{N_O\times N_I}$ and $C\in \mathbb{R}^{N_O\times d}$ is derived from KiUT that integrated symptom embeddings, probabilities, and a Graph Attention Network. $N_O$ denotes the number of symptoms extracted from the reports. \\
\noindent \textbf{IFR: Image Feature Refiner.} As shown in \Cref{fig:overview} (b), we introduce IFR that combines multi-class classification loss $\mathcal{L}_{cls-I}$ and a saliency map $S$, to enhance the extraction of fine-grained details from radiology images. Initially, we employ CheXBert~\cite{smit2020CheXBert} to extract labels from the training reports, designating ``blank" labels as negative and ``uncertain" labels as a unique category to refine the classification granularity. We calculate $\mathcal{L}_{cls-I}$ using Binary Cross Entropy (BCE) loss, defined per data point as $\mathcal{L}_{BCE}=-[y\log(\sigma(p))+(1-y)\log(1-\sigma(p))]$, where $y$ is the true label, $p$ is the predicted label, and $\sigma$ represents the sigmoid function. Since CheXBert extracts 14 classes for each paired image and report, the image classification loss is thus formulated as 
\begin{equation}
    \mathcal{L}_{cls-I}=\frac{1}{N}\sum_{i=1}^{N}[\frac{1}{14}\sum_{j=1}^{14}\mathcal{L}_{BCE}(y_{ij}, p_{ij})],
    \label{eq:BCE_Loss}
\end{equation} 
where $y_{ij}$ and $p_{ij}$ denote the ground truth and predicted labels for $k^{th}$ class of the $i^{th}$ example.
\begin{figure}
    \vspace{-3mm}
    \centering
    \includegraphics[width=0.9\linewidth]{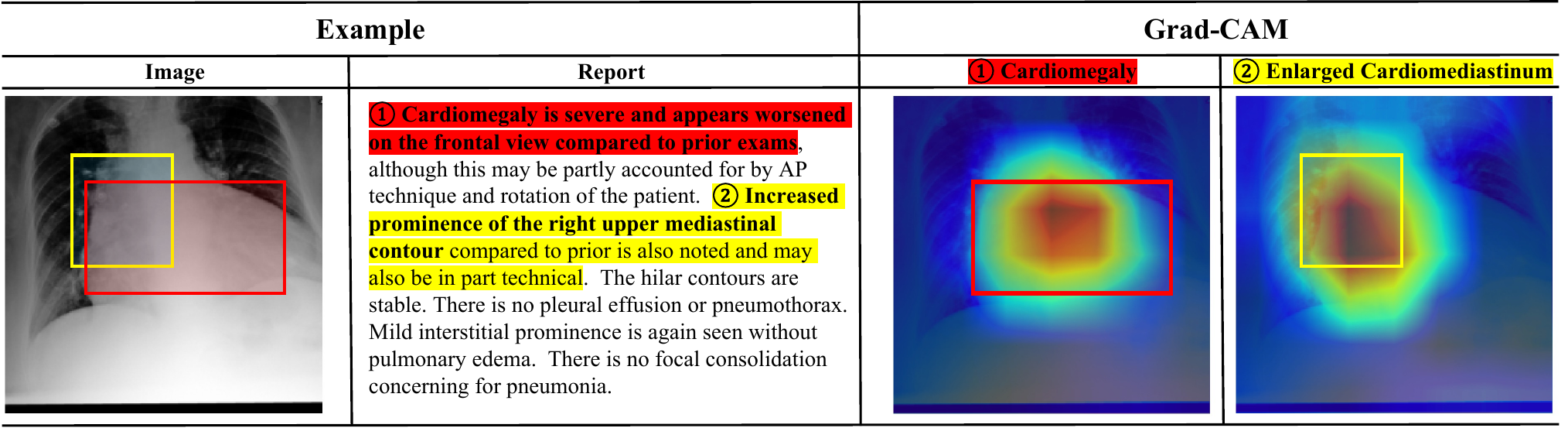}
    \caption{Illustration of the image-report pair and its corresponding Grad-CAM for the observations. The colored boxes indicate the ground truth abnormal regions and the colored weights present the corresponding attention in the Grad-CAM.}
    \label{fig:Grad-CAM}
    \vspace{-6mm}
\end{figure}
\par
Furthermore, as shown in \Cref{fig:Grad-CAM}, we leverage Grad-CAM obtained from TorchXRayVision~\cite{cohen2022torchxrayvision} as the saliency map $S$ to identify correspondences between image patches and symptoms (e.g., lung opacity).  As shown in \Cref{fig:overview}(b), the saliency map serves as prior information to merge visual features $V$ and symptom features $C$ to get refined image feature $\hat{C}$, thereby augmenting symptom representations. The detailed formulation is presented in the Appendix.\\
\noindent\textbf{TFR: Text Feature Refiner.}
We first construct the triplets by segmenting the training reports $R$ into segments $A=\{a_1, a_2, ..., a_{N_A}\}$ where $N_A$ is the length of segments. Each segment is annotated with a set of labels across 14 predefined classes using CheXBert~\cite{smit2020CheXBert}. For each class $k$, we construct triplet $t=\{a, p, n\}$ based on the annotation similarity. Specifically, a pair of sentences $(a_i, a_j)$ is considered positive if share the same annotation for class $k$, i.e., $label(a_i)_k=label(a_j)_k$ and negative if $label(a_i)_k\neq label(a_j)_k$. This process is repeated for each class, resulting in a set of triplets $T=\{t_1, t_2,...,t_{N_T}\}$, where $N_T$ is the number of triplets. To refine the text features, we fine-tune BERT~\cite{devlin2018bert} initialized from BERT-Base with multi-class classification loss $\mathcal{L}_{cls-T}$ on the training reports and triplet loss $\mathcal{L}_{triplet}$ on the triplets $T$. As shown in \Cref{fig:overview}(d), We employ the report $R$ and the extracted labels from CheXBert to calculate the multi-class classification loss $\mathcal{L}_{cls-T}$ as~\Cref{eq:BCE_Loss}. The triplet loss is formulated with the triplets $\mathcal{T}$ as
\begin{equation}
        \mathcal{L}_{triplet} = \frac{1}{|T|}\sum_{(a,p,n)\in T}\max(||f(a) - f(p)||^2 + ||f(a)-f(n)||^2+\beta, 0), 
\end{equation}
where $f(x) = \sigma(\textit{BERT}(x)W_t + b_t)$, $\beta\in[0, 1]$ is the margin parameter. During the training phase, BERT processes the entire report as input while it is applied to obtain refined text features $W_{1:t}$ of the generated report word at time step $t$ during the inference phase. \\
\noindent\textbf{CA: Contrastive Aligner.}
The refined visual and textual features, while beneficial in isolation, can lead to a divergence between the image and text representations. This misalignment may hinder the model's ability to establish accurate relationships between two modalities. Therefore, consistently aligning these modalities is a significant aspect for accurate report generation. As shown in \Cref{fig:overview} (e), we introduce an image-text contrastive loss as \cite{li2022blip,li2023blip} to enhance the consistency. The image-text contrastive loss, denoted as $\mathcal{L}_{itc}$, is formulated to align image and text representation such that their mutual information is maximized. We align the global image features $f_{image}=\textit{Avgpool}(V)$ and text features $f_{text}$, where $f_{text}=\textit{Mean}(W)$. The loss function is defined as:
\begin{equation}
    \begin{aligned}
        &\mathcal{L}_{itc} = \frac{1}{2N} \sum_{i=1}^{N} \left( l(f_{\text{image}_i}, f_{\text{text}}) + l(f_{\text{text}_i}, f_{\text{image}}) \right),
    \end{aligned}
\end{equation}
where $l(u, v) = -\log \left( \frac{\exp(\text{sim}(u_i, v_i) / \tau)}{\sum_{k=1}^{N} \exp(\text{sim}(u_i, v_k) / \tau)} \right)$, $\text{sim}(u,v)$ calculates the cosine similarity between vectors \textit{u} and \textit{v}, \textit{N} denotes the batch of \textit{N} samples.

\subsection{Training}
We first fine-tune the BERT to extract refined text features with the multi-classification loss $\mathcal{L}_{cls-T}$ and the triplet loss $\mathcal{L}_{triplet}$. Then we freeze the BERT model and proceed to the remaining part with the objective
\begin{equation}
    \mathcal{L} = \mathcal{L}_{CE} + \lambda_{cls-I} * \mathcal{L}_{cls-I} + \lambda_{itc} * \mathcal{L}_{itc},
\end{equation}
where $\lambda_{cls-I}$ and $\lambda_{itc}$ are hyper-parameter to control the loss contributions. We set $\lambda_{cls-I}=1$ and $\lambda_{itc}=0.1$ in our work.

\section{Experiments}
\subsection{Experimental Settings}
\textbf{Dataset.} We conduct our experiments on the widely-used public dataset, i.e., MIMIC-CXR dataset. The MIMIC-CXR dataset~\cite{johnson2019mimic} comprises 377,110 chest X-ray images and 227,835 reports from 64,588 patients. Following previous work~\cite{R2Gen,R2GenCMN,huang2023kiut}, we utilized the official split, allocating 70\% of the data for training, 10\% for validation, and 20\% for testing. This resulted in 368,960 samples in the training set, 2,991 in the validation set, and 5,159 in the test set. \\
\noindent \textbf{Evaluation Metrics.} To evaluate the performance, we employ the widely used natural language generation (NLG) metrics and clinical efficacy (CE) metrics. We utilize the evaluation protocol~\cite{chen2015microsoft} to calculate the NLG scores, including BLEU~\cite{bleu}, METEOR~\cite{banerjee2005meteor} and ROUGE-L~\cite{lin2004rouge}. In particular, BLEU and METEOR are proposed for machine translation evaluation. ROUGE-L is designed for evaluating the quality of summaries. For clinical efficacy, we apply CheXBert~\cite{smit2020CheXBert} to label the generated reports and calculate the precision, recall, and F1 through the predicted labels. \\
\noindent \textbf{Implementation Details.} 
The implementation details are based on previous works~\cite{R2Gen,huang2023kiut}. We begin by pre-training BERT~\cite{devlin2018bert} for 30 epochs using both triplet loss and classification loss. Subsequently, we freeze the BERT model and proceed to train the visual extractor and encoder-decoder architecture following the settings outlined in Huang et al.~\cite{huang2023kiut}.


\begin{table}[ht]
    \vspace{-3mm}
    \centering
    \caption{Results on the MIMIC-CXR benchmark, where $\uparrow$ denotes that higher values indicate better results. * denotes methods that introduce external annotations. $\dag$ indicates that the results are evaluated by us, ensuring consistent CE evaluation with other methods. The \textbf{best} and \underline{second best} results are highlighted.}
    \resizebox{0.9\textwidth}{!}{
    \begin{tabular}{c|cccccc|ccc}
    \hline
         \multirow{2}{*}{Methods} & \multicolumn{6}{c}{NLG $\uparrow$}  & \multicolumn{3}{|c}{CE $\uparrow$} \\  \cline{2-10}
                   & BLEU-1  & BLEU-2  & BLEU-3  & BLEU-4  & METEOR  & ROUGE-L  & P  & R  & F1  \\
        \hline
        \hline
        R2Gen~\cite{R2Gen}       & 0.353 & 0.218 & 0.145 & 0.103 & 0.142 & 0.277 & 0.333 & 0.273 & 0.276 \\
        R2GenCMN~\cite{R2GenCMN}    & 0.353 & 0.218 & 0.148 & 0.106 & 0.142 & 0.278 & 0.334 & 0.275 & 0.278 \\
        PPKED~\cite{ppked}       & 0.360 & 0.224 & 0.149 & 0.106 & 0.149 & 0.284 &   -   &   -   &   -   \\
        Contrastive~\cite{liu2021contrastive} & 0.350 & 0.219 & 0.152 & 0.109 & 0.151 & 0.283 & 0.352 & 0.298 & 0.303 \\
        AlignTrans~\cite{you2021aligntransformer}  & 0.378 & 0.235 & 0.156 & 0.112 & 0.158 & 0.283 &   -   &   -   &   -   \\
        MMTN~\cite{cao2023mmtn} & 0.379 & 0.238 & 0.159 & 0.116 & 0.161 & 0.283 & - & - & - \\
        METransformer~\cite{wang2023metransformer} & 0.386 & \underline{0.250} & \underline{0.169} & \textbf{0.124} & 0.152 & \underline{0.291} & 0.364 & 0.309 & 0.311\\
        KiUT~\cite{huang2023kiut} & \underline{0.393} & 0.243 & 0.159 & 0.113 & \underline{0.160} & 0.285 & \textbf{0.371} & \underline{0.318} & \underline{0.321} \\ 
        \hline
        Ours (FITA) & \textbf{0.411} & \textbf{0.262} & \textbf{0.173} & \underline{0.122} & \textbf{0.172} & \textbf{0.295} & \underline{0.365} &  \textbf{0.322}    & \textbf{0.326}     \\
        \hline 
        \hline
        RGRG$^{*}$~\cite{rgrg} & 0.373 & 0.249 & 0.175 & 0.126 & 0.168 & 0.264 & 0.461 & 0.475 & 0.447 \\
        PromptMRG$^{* \dag}$~\cite{jin2023promptmrg} & 0.357	& 0.219	& 0.146 & 0.103 & 0.145 & 0.272 & 0.469 & 0.359 & 0.379 \\
        \hline
        
    \end{tabular}}
    \label{tab:main_result}
    \vspace{-7mm}
\end{table}
\subsection{Results}
\textbf{Results on MIMIC-CXR Benchmark.} We first compare our method with a wide range of state-of-the-art radiology report generation models on the MIMIC-CXR. \Cref{tab:main_result} shows the comparison results on both NLG and CE metrics. The models include R2Gen~\cite{R2Gen}, PPKED~\cite{ppked}, KiUT~\cite{huang2023kiut}, et al. Our model outperforms the state-of-the-art method for the NLG metrics on MIMIC-CXR benchmark. 
\par
Our method exhibits lower performance on the CE metric when compared to RGRG~\cite{rgrg} and PromptMRG~\cite{jin2023promptmrg} since they introduced extra annotations provided by LLMs, CLIP~\cite{radford2021learning} or human doctors. RGRG adopted different data-split~\cite{wu2021chest2} which contains ground truth labels for abnormal visual regions. We evaluate PromptMRG under the same setting as previous works with the same decoder layer and report length. PromptMRG performs better on CE metric by utilizing LLMs (Vicuna 13B~\cite{zheng2024judging}) to assist in disease labeling and using CLIP to retrieve reports. However, it achieves lower NLG scores than our FITA, suggesting that PromptMRG is limited in generating reports that are consistent with the style of ground truth report. 

\par
Furthermore, since we adopt a structure similar to KiUT~\cite{huang2023kiut}, the comparison between FITA and KiUT indicates the effectiveness of fine-grained image-text alignment for radiology images and reports. Moreover, the better clinical efficacy that measures the accuracy of generated reports for abnormalities demonstrates that FITA can produce higher-quality reports for clinical abnormalities. \\
\noindent \textbf{Ablation Study.} To further explore the impact of each component in FITA, We conduct an ablation analysis to compare each component. The main results are shown in \Cref{tab:ablation_study}. For the baseline KiUT, we reproduce the results and report the CE metric obtained through CheXBert. The results firstly indicate that both IFR (\Cref{tab:ablation_study}(a)) and TFR (\Cref{tab:ablation_study}(b)) can improve the NLG and CE metrics. We hypothesize that the refined visual feature helps detect diseases and thus improves the accuracy of abnormal detection. TFR boosts the CE metric by effectively distinguishing sentence differences. The combination of IFR and TFR (\Cref{tab:ablation_study}(c)) further improves the NLG performance, which demonstrates the compatibility of the two modules. However, the observed decrease in the CE metric may result from misalignment between refined image and text features.
\begin{table}[htbp]
    \vspace{-3mm}
    \centering
    \caption{Ablation study of our method on the MIMIC-CXR dataset, where $\uparrow$ denotes that higher values indicate better results.}
    \resizebox{0.9\textwidth}{!}{
    \begin{tabular}{c|ccc|cccccc|ccc}
        \hline
        \multirow{2}{*}{Settings} & \multicolumn{3}{c}{Modules} & \multicolumn{6}{|c}{NLG $\uparrow$} & \multicolumn{3}{|c}{CE $\uparrow$} \\ 
        \cline{2-13}
        & IFR & TFR & CA & BLEU-1  & BLEU-2  & BLEU-3  & BLEU-4  & METEOR  & ROUGE-L  & P  & R  & F1 \\
        \hline
        \hline
        Baseline & \XSolidBrush & \XSolidBrush & \XSolidBrush & 0.393 & 0.243 & 0.159 & 0.113 & 0.160 & 0.285 & 0.376 & 0.256 & 0.262 \\
        (a) & \Checkmark & \XSolidBrush & \XSolidBrush & 0.396 & 0.251 & 0.167 & 0.119 & 0.164 & 0.294 & \textbf{0.378} & 0.309 & 0.317 \\
        (b) & \XSolidBrush & \Checkmark & \XSolidBrush & 0.398 &	0.253 &	0.167 &	0.118 &	0.168 &	0.294 &	0.367 &	0.301 &	0.302 \\
        (c) & \Checkmark & \Checkmark & \XSolidBrush & 0.406 & 0.260 & 0.173 & 0.123 & 0.171 & 0.297 & 0.365 & 0.286 & 0.291\\
        \hline
        (c) & \Checkmark & \Checkmark & \XSolidBrush & 0.406 & 0.260 & 0.173 & \textbf{0.123} & 0.171 & \textbf{0.297} & 0.365 & 0.286 & 0.291\\
        (d) - Ours & \Checkmark & \Checkmark & \Checkmark & \textbf{0.411} &	\textbf{0.262} & \textbf{0.173} &	0.122 &	\textbf{0.172} &	0.295 &	0.365 &	\textbf{0.322} & \textbf{0.326} \\
        \hline
    \end{tabular}}
    \label{tab:ablation_study}
    \vspace{-6mm}
\end{table}
\par
Next, we assess the effectiveness of Contrastive Aligner (CA) in addressing misalignment between refined image and text features. As illustrated in \Cref{tab:ablation_study}(d), CA can significantly enhance both the NLG and CE metrics. The increase in CE confirms that CA effectively mitigates the decrease in CE resulted from misalignment between refined image and text features, highlighting the importance of aligning these two distinct modalities.


\section{Conclusion}
In this paper, we present an effective approach termed FITA of capturing both the fine-grained details within radiological data and the crucial alignment between refined images and textual descriptions. The experiments and analyses on the MIMIC-CXR datasets verify the effectiveness of three modules in FITA: Image Feature Refiner, Text Feature Refiner and Contrastive Aligner. The ablation study reveals the decrease in CE metric resulted from misalignment of refined image and text features, which further highlights the significance of aligning these two distinct modalities. 
\vspace{-3mm}
%
%
%
\bibliographystyle{aaa}
\bibliography{ref}

\end{document}


\title{FITA: Fine-Grained Image-Text Alignment for Radiology Report Generation (Supplementary)}
\author{*}
\authorrunning{Anonymous}

\maketitle             
\section{Examples of the Constructed Triplets Dataset}

\begin{table}[htbp]
\caption{Examples of constructed triplets from the MIMIC-CXR dataset covering 13 observations (excluding "No Finding"). These observations are derived using CheXBert. For each triplet $t=(a, p,n)$, the anchor $a$ and positive sample $p$ represent the abnormal observation while the negative sample $n$ represents either a normal observation or does not specify the observation. "Enlarged Cardio." in the table stands for Enlarged Cardiomediastinum.}  

\resizebox{1.0\textwidth}{!}{
\begin{tabular}{l|l|l|l}
\hline
Symptoms   & Anchor & Positive & Negative  \\                                                                                                                                      \hline                                                                      
Enlarged Carido. & \begin{tabular}[c]{@{}l@{}}the cardiac mediastinal and hilar contours\\ are unchanged with similar aortic tortuosity\end{tabular}      & \begin{tabular}[c]{@{}l@{}}the cardiomediastinal silhouette is \\ stable noting mildly tortuous aorta\end{tabular}                                              & \begin{tabular}[c]{@{}l@{}}the lungs are clear the cardiomediastinal \\ silhouette and hila are normal\end{tabular}                                          \\
\hline
Cardiomegaly     & the heart size is borderline enlarged                                                                                                  & \begin{tabular}[c]{@{}l@{}}heart size remains mild to moderately \\ enlarged\end{tabular}                                                                       & the heart size is normal                                                                                                                                     \\
\hline
Lung Opacity     & \begin{tabular}[c]{@{}l@{}}a triangular opacity in the right lung apex \\ is new from prior examination\end{tabular}                   & \begin{tabular}[c]{@{}l@{}}no focal opacification is seen aside from \\ streaky left lower lung opacity suggesting \\ minor atelectasis\end{tabular}            & \begin{tabular}[c]{@{}l@{}}there is no airspace opacity worrisome \\ for pneumonia\end{tabular}                                                              \\
\hline
Lung Lesion      & \begin{tabular}[c]{@{}l@{}}numerous nodular opacities compatible the \\ patients metastatic disease are again appreciated\end{tabular} & \begin{tabular}[c]{@{}l@{}}subcentimeter nodular opacity in the \\ periphery of the left upper lobe also \\ appears similar to radiograph\end{tabular}          & there are no suspicious osseous lesions                                                                                                                      \\
\hline
Edema            & mild pulmonary vascular congestion is stable                                                                                           & \begin{tabular}[c]{@{}l@{}}compared with the prior radiograph \\ there is mild worsening of pulmonary \\ vascular congestion\end{tabular}                       & \begin{tabular}[c]{@{}l@{}}no signs of pneumonia or pulmonary \\ vascular congestion\end{tabular}                                                            \\
\hline
Consolidation    & \begin{tabular}[c]{@{}l@{}}there is a focal consolidation at the left \\ lung base adjacent to the lateral hemidiaphragm\end{tabular}  & \begin{tabular}[c]{@{}l@{}}additionally an area of consolidation has \\ developed within the left retrocardiac region\end{tabular}                              & \begin{tabular}[c]{@{}l@{}}there is no focal consolidation pleural \\ effusion or pneumothorax\end{tabular}                                                  \\
\hline
Pneumonia        & \begin{tabular}[c]{@{}l@{}}asymmetric opacity in the right middle\\ lobe is concerning for pneumonia\end{tabular}                      & \begin{tabular}[c]{@{}l@{}}as compared to the previous radiograph \\ there is a massive increase in extent and \\ severity of multifocal pneumonia\end{tabular} & \begin{tabular}[c]{@{}l@{}}there is no new focal opacity to \\ suggest pneumonia\end{tabular}                                                                \\
\hline
Atelectasis      & \begin{tabular}[c]{@{}l@{}}mild atelectasis is seen in the lung \\ bases without focal consolidation\end{tabular}                      & \begin{tabular}[c]{@{}l@{}}there is mild left base atelectasis seen on \\ the frontal view without clear correlate on \\ the lateral view\end{tabular}          & \begin{tabular}[c]{@{}l@{}}as compared to the previous radiograph \\ there is resolution of the pre-existing right \\ basal atelectasis\end{tabular}         \\
\hline
Pneumothorax     & no complications notably no pneumothorax                                                                                               & \begin{tabular}[c]{@{}l@{}}right apical pneumothorax with a diameter \\ of approximately 13 mm\end{tabular}                                                     & \begin{tabular}[c]{@{}l@{}}there is no focal consolidation pleural \\ effusion or pneumothorax\end{tabular}                                                  \\
\hline
Pleural Effusion & \begin{tabular}[c]{@{}l@{}}small pleural effusion in the right \\ middle fissure is new\end{tabular}                                   & tiny bilateral pleural effusions are new since                                                                                                                  & no pleural effusion or pneumothorax is present                                                                                                               \\
\hline
Pleural Other    & there is bilateral apical pleural thickening                                                                                           & there is minimal biapical pleural thickening                                                                                                                    & lungs are clear                                                                                                                                              \\
\hline
Fracture         & \begin{tabular}[c]{@{}l@{}}remote left-sided rib fractures are \\ also re-demonstrated\end{tabular}                                    & \begin{tabular}[c]{@{}l@{}}again seen are multiple clips projecting over \\ the left breast and remote left-sided rib fractures\end{tabular}                    & \begin{tabular}[c]{@{}l@{}}osseous and soft tissue structures are \\ unremarkable specifically there is no \\ visualized displaced rib fracture\end{tabular} \\
\hline
Support Devices  & \begin{tabular}[c]{@{}l@{}}the tip appears to project over the \\ azygous vein at the level of the upper svc\end{tabular}              & \begin{tabular}[c]{@{}l@{}}enteric tube courses below the level of \\ the diaphragm\end{tabular}                                                                & \begin{tabular}[c]{@{}l@{}}there has been removal of the right ij \\ central venous line\end{tabular} \\
\hline
\end{tabular}}  
\vspace{-6mm}
\end{table}


\section{Detailed formulation for Image Feature Refiner}
\begin{equation}
\centering
\begin{aligned}
    &\hat{v}_{i} = MHA_s(c_i, V, s_i, \alpha) = (\frac{c_i W_s^Q(VW_s^K)^T}{\sqrt{d_k}} + \alpha s_i)VW_s^V \\
    &\hat{c}_i = c_i + MHA(v_i, C) = c_i + (\frac{v_i W^Q(CW^K)^T}{\sqrt{d_k}})CW^V,
\end{aligned}
\end{equation}
where $v_i$, $c_i$, $s_i$ is the $i^{th}$ element in $V$, $C$, and $S$, $\alpha$ is a hyper-parameter to control the contribution of saliency map $S$.